\pdfoutput=1
\documentclass[11pt]{article}

\usepackage[final]{acl}
\usepackage{booktabs}
\usepackage{times}
\usepackage{makecell}
\usepackage{balance}
\usepackage{multirow}
\usepackage{flushend}
\usepackage{amsmath}
\usepackage{latexsym}
\usepackage{enumitem}
\usepackage{flushend}
\usepackage{etoolbox}
\usepackage{indentfirst}
\usepackage{flafter}
\usepackage{placeins}
\usepackage{dblfloatfix}

\usepackage{titlesec}

\titlespacing*{\section}
{0pt}{*1.2}{*0.8}

\titlespacing*{\subsection}
{0pt}{*1}{*0.6}

\usepackage{listings}
\usepackage[T1]{fontenc}

\usepackage[utf8]{inputenc}

\usepackage{microtype}

\usepackage{inconsolata}

\usepackage{graphicx}

\title{5ting at SemEval-2026 Task 8: Strong End-to-End Multi-Turn RAG via LLM-Based Reranking and Faithfulness Control}

\author{
Thien-Qua T. Nguyen$^{1,2}$\quad
Chi Hoang$^{1,2}$ \quad
Nguyen Tran$^{1,2}$ \\
\textbf{Tri Le$^{1,2}$} \quad
\textbf{Khanh Truong$^{1,2}$} \quad
\textbf{Chinh Trong Nguyen$^{1,2}$} \\
$^{1}$University of Information Technology, Ho Chi Minh City, Vietnam \\
$^{2}$Vietnam National University Ho Chi Minh City, Ho Chi Minh City, Vietnam \\
\texttt{\{quantt.20, chihqqc.20, nguyenttb.20, triln.20, khanhtpb.20\}@grad.uit.edu.vn} \\
\texttt{chinhnt@uit.edu.vn}
}

\begin{document}
    \maketitle
\vspace{-1.2em}
\begin{abstract}
We introduce 5ting, our system for the \mbox{SemEval–2026} Task 8 (MTRAGEval), which evaluates multi-turn Retrieval-Augmented Generation (RAG) systems. Multi-turn RAG involves context drift, underspecification, and hallucination risk. Our system combines BGE-M3 dense retrieval with FAISS indexing, dual-query merged retrieval, and LLM-based reranking, followed by role-separated generation constrained to retrieved evidence.
The retriever achieved $nDCG@5 = 0.4719$ in Task~A, while the end-to-end system ranked in Task~C with a harmonic score of $0.5597$ and $RL_F = 0.7692$.

\end{abstract}

\section{Introduction}

RAG grounds LLM outputs in external knowledge to improve factual accuracy \citet{gao2024ragretrievalaugmentedgenerationlarge}. In multi-turn settings, challenges such as context dependency, underspecification, and conversational drift complicate both retrieval and generation \cite{owoicho2023exploiting}. To systematically evaluate these phenomena, \citet{rosenthal2026mtragunbenchmarkopenchallenges} introduced MTRAG-UN, targeting unanswerable, underspecified, and non-standalone questions, and showing that even state-of-the-art systems struggle in multi-turn scenarios.

SemEval-2026 Task 8 (MTRAGEval) \citet{Rosenthal2026MTRAGEval} evaluates multi-turn RAG across four domains: ClapNQ, FiQA, Cloud, and Govt. The final query may depend on prior dialogue context and contain implicit or ambiguous references, increasing retrieval difficulty and hallucination risk during generation.

We participate in Subtask A (top-10 passage retrieval) and Subtask C (grounded answer generation from top-5 passages). We present 5ting, a modular multi-turn RAG system combining dense retrieval, FAISS indexing, LLM-based reranking, and evidence-constrained generation. Rather than proposing a new retriever, we focus on reranking and prompt design.

Our main contributions are as follows:
\vspace{-4pt}
\begin{itemize}
    \setlength{\itemsep}{2pt}
    \setlength{\parskip}{0pt}
    \setlength{\topsep}{2pt}
    \item We design an LLM-based reranking mechanism that refines dense retrieval results in multi-turn settings.
    \item We introduce a faithfulness-oriented prompting strategy that constrains generation to retrieved evidence.
    \item We provide empirical analysis demonstrating retrieval--generation synergy.
\end{itemize}
\vspace{-4pt}

The remainder of the paper is organized as follows: Section 2 discusses multi-turn RAG challenges, Section 3 presents the system architecture, Section 4 describes the experimental setup, Section 5 reports results and ablations, and Sections 6–8 provide error analysis, limitations, and conclusions.

\section{Task Definition and Challenges}

The MTRAGEval benchmark evaluates retrieval and end-to-end RAG on 842 multi-turn dialogues across four domains. Subtask~A retrieves top-10 passages from a corpus of 78,170 documents and is evaluated using $nDCG@5$ (primary), $nDCG@{1,3,10}$, and $Recall@{1,3,5,10}$. Subtask~C requires grounded answer generation from the top-5 passages and is evaluated by the harmonic mean of $RB_{alg}$ (reference-based algorithmic relevance), $RL_F$ (RAGAS Faithfulness LLM Judge), and $RB_{llm}$ (LLM-as-judge relevance).

Compared to single-turn benchmarks, multi-turn dialogue introduces additional challenges. MTRAG-UN explicitly catalogs these phenomena, motivating our system design.

\begin{itemize}
    \setlength{\itemsep}{2pt}
    \setlength{\parskip}{0pt}
    \setlength{\topsep}{2pt}
    \item Context Dependency: Final queries often rely on prior turns and may be underspecified, requiring cross-turn interpretation.
    
    \item Retrieval--Generation Alignment: Strong retrieval does not ensure faithful generation; outputs must be explicitly grounded in retrieved evidence.
    
    \item Hallucination Control: Models may introduce unsupported information, requiring explicit grounding constraints.
    
    \item Non-Standalone and Unanswerable Cases: MTRAG-UN \citet{rosenthal2026mtragunbenchmarkopenchallenges} highlights ambiguous or evidence-absent queries that challenge multi-turn RAG systems.
\end{itemize}

Multi-turn RAG requires a tightly integrated pipeline: retrieval must balance surface-level matching with contextual understanding, while generation must remain grounded in retrieved evidence. Prior work shows that query rewriting and LLM-based reranking improve conversational retrieval (\citet{vakulenko2021question}; \citet{sun2023chatgpt}; \citet{ma2023zeroshot}), and grounding generation in retrieved evidence reduces hallucination and improves faithfulness \citet{gao2024ragretrievalaugmentedgenerationlarge}.

We address these challenges through three components: dual-query merged retrieval, LLM-based reranking, and role-separated prompting. Together, they improve precision \citet{liu2024chatqa} and enforce faithful generation in multi-turn settings.

\section{System Overview}

\begin{figure*}[t]
    \centering
    \includegraphics[width=0.9\textwidth]{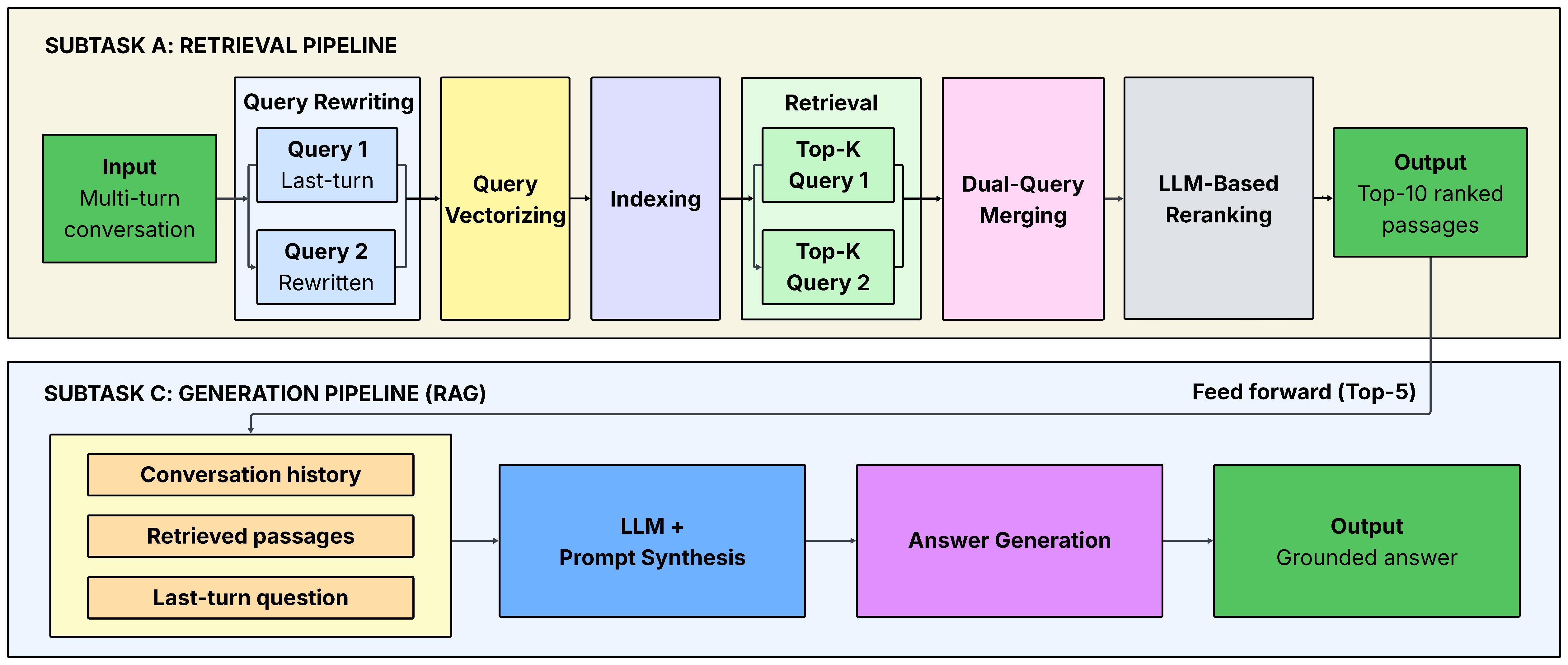}
    \caption{Overview of the proposed multi-turn RAG pipeline.}
    \label{fig:pipeline}
\end{figure*}

Our system consists of two main pipelines: a retrieval pipeline (Subtask A) and a generation pipeline (Subtask C), where the latter consumes the top-5 outputs from the former. Figure \ref{fig:pipeline} illustrates the overall architecture of our approach.

\subsection{Query Rewriting}

In multi-turn dialogue, the final query is often underspecified—containing unresolved coreferences, ellipsis, or implicit dependencies on prior turns—making direct retrieval unreliable. We use GPT-4o-mini with temperature set to 0.3 to rewrite the last-turn query into a self-contained, context-resolved form by incorporating the full conversation history (\citet{vakulenko2021question}; \cite{ma2023zeroshot}). The rewritten query is then used for retrieval and reranking, ensuring downstream components operate on an unambiguous representation of the user's information need.

\subsection{Query Vectorizing, Indexing and Retrieval}

Query Vectorizing: We use BAAI/bge-m3 \citet{chen2024bge} as our dense encoder, producing 1024-dimensional L2-normalized embeddings that support input sequences up to 8192 tokens. Both the last-turn and rewritten queries are encoded independently. Large-context dense encoders trained on diverse supervised data provide strong domain-general representations and robust cross-domain generalization \citet{wang-etal-2024-improving-text}.

For indexing, we adopt FAISS IndexFlatIP \citet{johnson2021billion}, which performs inner-product search over normalized vectors, equivalent to cosine similarity. We construct one index per domain and store document identifiers and associated metadata for efficient lookup. At query time, we encode the query using the same model and retrieve the top-$k$ passages ranked by similarity.

Retrieval: At inference time, the last-turn and rewritten queries independently search their respective domain indices, each retrieving the top-10 passages. Using FAISS enables efficient similarity search with minimal computational overhead. The two retrieval streams run in parallel to increase recall coverage, providing a diverse candidate set before the merging and LLM-based reranking stages.

\subsection{Dual-Query Merging}
We maintain two parallel retrieval streams: one using the last-turn query (surface-level lexical match) and one using the rewritten query (contextual semantic match). For each task, we perform two separate retrieval passes, one per query type, each returning the top-10 candidate sets. We then merge the two result sets. For passages appearing in both sets, we keep the maximum similarity score and the merged set is sorted by score in descending order, yielding approximately 10–20 unique candidate passages.

This dual-query strategy addresses a fundamental tension in multi-turn retrieval: the last-turn query may match passages that contain the exact entities mentioned by the user, while the rewrite query may retrieve passages that are semantically relevant to the broader conversational context. The merging of both captures the strengths of each.

\subsection{LLM-Based Reranking}

We apply a listwise reranking step using GPT-4o-mini with temperature is 0.3 to refine the ranking of the merged candidates. Specifically, we adopt a listwise strategy to optimize the ranking from a global perspective, allowing the model to jointly consider all candidate passages and produce a more coherent overall ordering, albeit at a higher computational cost than setwise reranking. The reranker receives the rewrite query (which captures conversational context) and the text of all candidate passages and outputs a JSON-structured ranking.

The system prompt instructs the model: "You are a document re-ranker. Given a question and a set of document chunks, rank them by relevance. Return a JSON object with the key 'order' containing an array of IDs sorted from most to least relevant."

Rationale for using the rewrite query for reranking: The rewrite query is a self-contained representation of the user's information need \citet{zhuang2024setwise}; \citet{sun2023chatgpt}, making it more suitable for the LLM reranker, which does not have access to the conversation history and has been shown to substantially improve ranking precision in multi-step retrieval pipelines (\citet{pradeep2023rankzephyreffectiverobustzeroshot}).

\subsection{The LLM, Prompt Synthesis and Answer Generation}

This module serves as the core reasoning interface within the RAG pipeline. It integrates the retrieved evidence and the dialog context into a structured prompt that is then passed to the answer-generation model. The prompt design explicitly distinguishes between conversational context, which is used only to resolve intent, and retrieved passages, which serve as the sole factual source.

For Subtask C, we take the top-5 retrieved passages from our Subtask A output and generate an answer using GPT-4o-mini. The prompt is structured with clear role separation:

\vspace{-4pt}
\begin{itemize}    
    \setlength{\itemsep}{2pt}
    \setlength{\parskip}{0pt}
    \setlength{\topsep}{2pt}
    \item Conversation History (all turns except the last): used only for understanding user intent and resolving references; explicitly stated to not be a source of factual information.
    \item Retrieved Passages: designated as the sole source of facts.
    \item Last-Turn Question: the specific question to answer.
\end{itemize}
\vspace{-4pt}

This design reflects the principle that in a RAG system, the conversation history provides pragmatic context (what the user is talking about, what has already been discussed, etc) while the retrieved passages provide epistemic grounding (the factual basis for the answer). Evidence-grounded prompting of this form has been shown to reduce hallucination and improve factual fidelity in RAG systems.

\section{Experimental Setup}

\subsection{Data}
MTRAGEval builds on the MTRAG dataset \citet{mtrag}, comprising 110 conversations (avg. 7.7 turns) across four domains, yielding 842 evaluation tasks. MTRAG-UN \citet{rosenthal2026mtragunbenchmarkopenchallenges} extends this with 666 additional tasks across six domains. The corpus contains 78,170 documents and 366,479 passages (512 tokens, 100-token overlap). Retrieval uses three query variants (question, lastturn, rewrite), and Subtask~C follows the full RAG setting with generation from the top-5 passages.

\subsection{Evaluation Metrics}
All experiments were conducted using the official SemEval--2026 Task~8 evaluation framework. Retrieval performance was evaluated using $nDCG@5$ along with $nDCG@{1, 3, 10}$ and $Recall@{1, 3, 5, 10}$ for Task~A. End-to-end performance for Task~C was measured using the harmonic mean of $RB_{\text{alg}}$, $RL_F$, and $RB_{\text{llm}}$.

We follow the official evaluation protocol and exclude underspecified examples as required for reporting final scores. 

\subsection{LLM Model and Configuration}
We use the publicly available GPT-4o-mini model without task-specific fine-tuning for query rewriting, reranking, and answer generation, configuring only the decoding hyperparameters for generation with nucleus sampling ($\text{top-p} = 0.9$) and a low temperature ($T=0.1$) to promote stable and faithful generation. The $top-p$ constraint limits decoding to high-probability tokens, while the low temperature sharpens the distribution to encourage near-deterministic outputs. This setup enhances consistency and factual grounding, making it suitable for retrieval-augmented generation, where minimizing hallucination is critical. For reranking, we use listwise prompting over the full merged candidate set and do not apply a sliding-window mechanism.

\section{Results and Analysis}
\FloatBarrier
\subsection{Official Test Results}

Subtask A: Our $nDCG@5$ reaches 98.4\% of the ELSER baseline (gap = 0.0076) as shown in Table \ref{tab:retrieval_dev}, despite using a general-purpose dense encoder versus a task-specific learned sparse retriever used during benchmark creation. The dev-to-test increase from 0.439 to 0.4719 in nDCG@5 occurred without any changes to the pipeline settings. We attribute the difference to dataset-specific variation in domain composition and query difficulty, together with the sensitivity of nDCG@5 to top-ranked passages.

Subtask C: Our system surpasses all organizer baselines by +4.3\% and achieves 95\% of the top score (Table \ref{tab:generation_full_rag}), demonstrating that a well-engineered prompt can outperform larger reasoning-capable models. The high $RL_F$ (0.7692) confirms strong lexical faithfulness attributable to our grounding instructions.

\begin{table}[!b]
\small
\centering
\renewcommand{\arraystretch}{1.1}
\setlength{\tabcolsep}{6pt}
\begin{tabular}{l l c c}
\toprule
\textbf{System} & \textbf{Type} & \textbf{nDCG@5} & \textbf{R@5} \\
\midrule
BM25                & Lexical      & 0.22  & 0.25 \\
BGE-base 1.5        & Dense        & 0.34  & 0.37 \\
Ours w/o reranker   & Dense        & 0.373 & 0.37 \\
ELSER               & Sparse       & \textbf{0.48}  & \textbf{0.52} \\
Ours + reranker     & Dense+LLM    & 0.439 & 0.44 \\
\midrule
Our (test phase) & Dense+LLM & 0.4719 & -- \\

\bottomrule
\end{tabular}
\caption{Retrieval performance comparison on the dev set (Query Rewrite, 4-domain average). Baseline values are from the official MT RAG Benchmark.}
\label{tab:retrieval_dev}
\end{table}

\begin{table}[!t]
\small
\renewcommand{\arraystretch}{1.1}
\setlength{\tabcolsep}{6pt}
\centering
\begin{tabular}{lcccc}
\toprule
Subtask A & ClapNQ & Cloud & FiQA & Govt
\\
\midrule
nDCG@5 & 0.5112 & 0.4019 & 0.3872 & 0.4569 \\
Recall@5 & 0.5327 & 0.4237 & 0.3919 & 0.5027 \\
\bottomrule
\end{tabular}
\caption{Per-domain retrieval performance of our system on the four MTRAG (dev set, rewrite + rerank).}
\label{tab:per_domain_analize}
\end{table}

\begin{table}[!t]
\small
\renewcommand{\arraystretch}{1.1}
\setlength{\tabcolsep}{6pt}
\centering
\begin{tabular}{lccc}
\toprule
\textbf{Query Type} 
& \makecell{\textbf{nDCG@5} \\ \textbf{(Base)}} 
& \makecell{\textbf{nDCG@5} \\ \textbf{(+Reranker)}} 
& \makecell{\textbf{$\Delta$ rel} \\ \textbf{(\%)}} \\
\midrule
questions & 0.2066 & 0.2192 & +6.1  \\
lastturn  & 0.3454 & 0.4029 & +16.6 \\
rewrite   & \textbf{0.3732} & \textbf{0.4393} & +17.7 \\
\bottomrule
\end{tabular}
\caption{Ablation study on the effect of LLM reranking (dev set, 4-domain average).}
\label{tab:ablation_reranking}
\end{table}

\begin{table}[!t]
\small
\centering
\renewcommand{\arraystretch}{1.15}
\begin{tabular}{lccc}
\toprule
\textbf{Configuration} & \textbf{nDCG@1} & \textbf{nDCG@5} & \textbf{R@10} \\
\midrule
BGE-M3 dense & 0.2564 & 0.3454 & 0.4641 \\
+ Query rewrite & 0.3590 & 0.3732 & 0.5312 \\
+ LLM reranking & \textbf{0.4546} & \textbf{0.4393} & \textbf{0.5588} \\
\bottomrule
\end{tabular}
\caption{Cumulative ablation results (dev set, 4-domain average).}
\label{tab:cumulative_ablation}
\end{table}

Notably, the performance gap between Subtask A and Subtask C reveals a consistent rank asymmetry, where moderate retrieval quality is amplified at the generation stage. This suggests a structural property of multi-turn RAG systems: downstream reranking and grounded generation emphasize high-precision evidence, making top-ranked context more critical than overall recall. As a result, effective generation-side faithfulness control can partially compensate for retrieval limitations.

Table \ref{tab:per_domain_analize} shows clear domain-specific variation. ClapNQ achieves the best performance ($nDCG@5 = 0.5112$), indicating strong alignment with dense semantic retrieval, while FiQA is the most challenging ($nDCG@5 = 0.3872$), likely due to domain-specific terminology and higher query ambiguity. Cloud and Govt exhibit intermediate results.

The relatively small gap between $nDCG@5$ and $Recall@5$ across domains suggests that reranking mainly improves the ordering of relevant passages rather than recall. Overall, these results confirm that domain characteristics, particularly in FiQA, remain a key bottleneck for retrieval performance.

\begin{table*}[t]
\centering
\small
\renewcommand{\arraystretch}{1.1}
\setlength{\tabcolsep}{6pt}
\begin{tabular}{c l c c c c c}
\toprule
\textbf{Phase} & \textbf{Model} & \textbf{Size} & \textbf{RB\_llm} & \textbf{RL\_F} & \textbf{RB\_alg} & \textbf{Combined} \\
\midrule
\multirow{9}{*}{dev}
& Llama 3.1 Instruct   & 8B    & 0.54   & 0.53   & 0.34     & 0.45      \\
& Llama 3.1 Instruct   & 70B   & 0.59   & 0.64   & \textbf{0.39}     & 0.52      \\
& Llama 3.1 Instruct   & 405B  & 0.63   & 0.65   & \textbf{0.39}     & 0.53      \\
& Mixtral 8$\times$22B & 141B  & 0.61   & 0.56   & 0.35     & 0.48      \\
& Command-R+           & 104B  & 0.59   & 0.66   & 0.38     & 0.51      \\
& Qwen 2.5             & 72B   & 0.65   & 0.64   & 0.37     & 0.52      \\
& GPT-4o-mini          & --    & 0.64   & 0.64   & 0.37     & 0.52      \\
& GPT-4o               & --    & 0.66   & 0.65   & 0.38     & 0.53      \\
& Ours                 & --    & \textbf{0.67} & \textbf{0.77} & 0.38 & \textbf{0.55} \\
\midrule
\multirow{2}{*}{test}
& qwen-30b-a3b-thinking & 30B   & --     & --     & --     & 0.5366  \\
& Ours                  & --    &  \textbf{0.6784}     & \textbf{0.7692}     & \textbf{0.3867} & \textbf{0.5597} \\
\bottomrule
\end{tabular}
\caption{Generation baselines and our system under the full RAG setting on the dev and test phases. Bold indicates the best score within each phase.}
\label{tab:generation_full_rag}
\end{table*}

\begin{table*}[!t]
\begin{tabular}{l c p{9cm}}
\toprule
\textbf{Failure Type} & \textbf{Count} & \textbf{Example} \\
\midrule
Coreference not resolved & 12 (40\%) & ``What about the second option?'' — the model cannot determine which ``option'' is referenced \\
Topic shift & 7 (23\%) & The conversation shifts domains within the same session \\
Long-range dependency & 5 (17\%) & The information need depends on context from 3+ turns ago \\
Rare terminology & 4 (13\%) & Domain-specific jargon not well represented in embedding \\
Annotation ambiguity & 2 (7\%) & Multiple valid passages could answer the query \\
\bottomrule
\end{tabular}
\caption{Error analysis by failure type.}
\label{tab:error_analysis}
\end{table*}

\subsection{Baseline Comparison}

Our base retriever already outperforms BGE-base 1.5 by +9.8\% (Table \ref{tab:retrieval_dev}), validating the BGE-M3 upgrade. With reranking, we reach 91.5\% of ELSER's $nDCG@5$. Our $nDCG@1$ (0.455) nearly matches ELSER (0.46), showing the reranker's effectiveness at top-1 precision.

In Table \ref{tab:generation_full_rag}, our $RL_F$ (0.7692) exceeds the best baseline $RL_F$ (GPT-4o: 0.66) by +16.5\%, demonstrating that prompt design contributes more to lexical faithfulness than model scale. Our $RB_{llm}$ (0.6784) matches GPT-4o's (0.68), indicating comparable answer quality despite a smaller model.

\vspace{4pt}

Two factors explain the result:

\begin{itemize}
    \setlength{\itemsep}{2pt}
    \setlength{\parskip}{0pt}
    \setlength{\topsep}{2pt}
    \item Precision over recall in context: We feed only top-5 reranked passages. The LLM reranker's +17.7\% $nDCG@5$ improvement (Table \ref{tab:ablation_reranking}) ensures the top passage is highly relevant, giving the generator concentrated evidence rather than noisy context. 
    \item Anti-hallucination Prompting: We separate dialogue history (intent) from retrieved passages (facts) to prevent the model from misinterpreting prior turns as factual evidence, a common failure mode in multi-turn RAG.
\end{itemize}

\vspace{-4pt}
The harmonic mean penalizes metric imbalance, favoring systems with consistent performance. Our relatively balanced scores yield a stronger combined result. The main bottleneck is $RB_{alg}$ (0.3867), which reflects reference alignment. Improving retrieval recall would likely increase reference-aligned content and boost this component.

\subsection{Ablation: Effect of LLM Reranking}

The reranker and query rewriting are complementary: the reranker's gain is largest on rewrite (+17.7\%) as shown in Table \ref{tab:ablation_reranking}, indicating that it refines ordering after query rewriting improves recall, demonstrating the reranker's effectiveness at promoting the most relevant passage.

\subsection{Component Contribution}

Table \ref{tab:cumulative_ablation} highlights the effectiveness and complementarity of the two components. Query rewriting primarily improves recall (+14.5\% $R@10$) and early precision (+40.0\% $nDCG@1$), expanding semantic coverage for multi-turn queries. LLM reranking then substantially refines ranking quality, contributing the largest gain at $nDCG@5$ (+17.7\%) and delivering 2.3× the standalone improvement of rewriting. Overall, the full pipeline achieves a +27.2\% relative $nDCG@5$ improvement, demonstrating the strong precision–recall synergy of the proposed approach.

\section{Error Analysis}
We analyze representative failure cases to characterize system limitations and manually examined 30 retrieval failures (cases where $Recall@10$ < 0.25) on the dev set, shown in Table \ref{tab:error_analysis}. Coreference dominance highlights dependence on upstream query rewriting quality. These failure modes align closely with the open challenges catalogued in MTRAG-UN \citet{rosenthal2026mtragunbenchmarkopenchallenges}, which finds that unanswerable, underspecified, and non-standalone questions remain particularly difficult for current RAG systems, confirming the generality of our error patterns. For generation, low-scoring cases stem from insufficient retrieval context, over-extraction, and occasional language mismatch.

For Subtask C, we examined cases where the LLM-as-judge score was below 0.3 (lowest quartile). Common issues include:

\vspace{-4pt}
\begin{itemize}
    \setlength{\itemsep}{2pt}
    \setlength{\parskip}{0pt}
    \setlength{\topsep}{2pt}
    \item Insufficient context: When retrieval fails (low recall), the generator either produces generic answers or acknowledges inability to answer.
    \vspace{4pt}
    \item Over-extraction: For some queries, the model copies large portions of retrieved passages instead of synthesizing concise answers.
\end{itemize}
\vspace{-4pt}

\section{Limitations}
Our system depends on commercial LLM APIs for both reranking and generation, which limits full reproducibility.  The reranking step adds latency (1–2 seconds per query) that may be prohibitive for real-time applications.  Our error analysis is based on a manual sample and may not generalize to all failure modes.  Additionally, we did not explore fine-tuning the embedding model on the target task data, which could potentially improve retrieval performance. We also did not evaluate hybrid retrieval with BGE-M3 or open-source rerankers such as RankZephyr, both of which remain promising directions for future work.

\section{Conclusion}

We presented Team 5ting’s system for SemEval-2026 Task 8, ranking 14/38 in Subtask A and 3/29 in Subtask C. Our results show that LLM-based reranking is the most impactful component, improving $nDCG@5$ by +17.7\%. A general-purpose dense retriever with reranking achieves 98.4\% of the ELSER baseline, demonstrating strong efficiency without specialized sparse models. More importantly, the Task A/C rank asymmetry suggests that generation-side grounding can compensate for moderate retrieval performance, highlighting the central role of faithfulness control in multi-turn RAG. Additionally, error analysis highlights challenges in context alignment and evidence aggregation, motivating future work on hybrid retrieval, open-source rerankers, and embedding fine-tuning.

\bibliography{custom}

\appendix
\section{System Prompt for Answer Generation (Subsection 3.5)}

To ensure transparency and reproducibility, we provide the full system
prompt used in the generation stage (Subtask C). This prompt
operationalizes our role-separated prompting strategy described in
Section~3.6 and explicitly enforces evidence-grounded generation.

The design clearly distinguishes three functional roles:
(1) conversation history for contextual interpretation only,
(2) retrieved passages as the sole factual authority, and
(3) the last-turn question as the target query.
The prompt also contains strict anti-hallucination instructions that
prohibit the model from introducing unsupported information.

\bigskip
\noindent\textbf{System prompt} (an example used for Subtask C)

\medskip
"""You are a precise and faithful question-answering assistant operating
within a Retrieval-Augmented Generation (RAG) pipeline. Your sole task
is to answer the user's final question using ONLY the factual information
present in the Retrieved Passages provided below.

\medskip
\noindent ROLE DEFINITIONS — READ CAREFULLY BEFORE ANSWERING
\medskip

\medskip
\noindent [ROLE 1 — CONVERSATION HISTORY: For context resolution ONLY]

\vspace{4pt}
The conversation history is provided exclusively to help you:
\begin{itemize}
\item Identify what topic the conversation concerns.
\item Resolve coreferences and pronouns (e.g., ``it'', ``that option'', ``the previous method'').
\item Identify what has already been discussed or clarified in prior turns.
\end{itemize}

\textit{CRITICAL:} The conversation history is NOT a source of factual information.
Do NOT extract, infer, or cite any facts from it. Treat it as pragmatic context only.

\bigskip
\noindent [ROLE 2 — RETRIEVED PASSAGES: The SOLE source of facts]

The retrieved passages below represent the only authoritative factual basis for your answer.
Every factual claim in your response MUST be directly supported by at least one of these passages.
If the passages do not contain sufficient information to answer the question, you must state that explicitly.

\bigskip
\noindent [ROLE 3 — LAST-TURN QUESTION: The specific question to answer]

Answer the question directly, concisely, and faithfully, grounded in the retrieved passages.

\bigskip
\noindent CONVERSATION HISTORY (intent and reference resolution only — NOT a factual source)

\smallskip
\textit{User:} What kind of equipment are members of the California Conservation Corps likely to learn how to use?

\textit{Agent:} Members of the California Conservation Corps (CCC) are likely to ...

\textit{User:} I also want to know if California Conservation Corps provides any uniforms when I join.

\textit{Agent:} Yes, when you join the California Conservation Corps (CCC), you will receive a uniform. The CCC uniform not only makes you look good, but also protects you. It is included as part of your membership. The uniform includes a soft cap, shirt, pants, hard hat, steel-toe boots, and gloves.

\bigskip
\noindent LAST-TURN QUESTION

\smallskip
\textit{User:} thank you. What does CCC’s Core Training program is for?
What course are offered during the training?

\bigskip
\noindent RETRIEVED PASSAGES (the sole factual basis for your answer)

\smallskip
\textit{text1:} CCC Core Training | California Conservation Corps  
Core Training includes the following courses:  
COMET: All applicants must complete ...

\smallskip
\textit{text2:} Training and Career Pathways | California Conservation Corps  
When it’s time to put the training into practice, the work they are ...

\smallskip
\textit{text3:} CCC Core Training | California Conservation Corps  
CCC Core Training | California Conservation Corps ...

\smallskip
\textit{text4:} Life in the Corps | California Conservation Corps  
In the CCC, crews will travel up to an hour or so to reach a project site ...

\smallskip
\textit{text5:} Training and Career Pathways | California Conservation Corps  
Training and Career Pathways | California Conservation ...

\bigskip
\noindent ANSWER GENERATION GUIDELINES

\begin{enumerate}
\item Ground every factual claim EXCLUSIVELY in the Retrieved Passages. Do not use parametric world knowledge.
\item If the passages are insufficient to answer the question, respond:  
``Based on the available retrieved information, I cannot provide a complete answer to this question.''
\item Do NOT speculate, hallucinate, or introduce any information absent from the passages.
\item Use the Conversation History only to resolve coreferences and clarify the question’s intent.
\item Synthesize information across passages coherently rather than quoting them verbatim.
\item Be concise and direct. Avoid unnecessary preamble or meta-commentary.
\item If the question is underspecified or ambiguous, acknowledge the ambiguity briefly before answering."""
\end{enumerate}

\bigskip

\end{document}